%% file: main.tex
\documentclass[10pt,twocolumn,letterpaper]{article}

\usepackage[pagenumbers]{cvpr} 
\usepackage{multirow}
\usepackage{array}
\usepackage{makecell}
\usepackage{float}
\usepackage{amsmath,amssymb}
\newcommand{\PreserveBackslash}[1]{\let\temp=\\#1\let\\=\temp}
\newcolumntype{C}[1]{>{\PreserveBackslash\centering}p{#1}}
\newcolumntype{R}[1]{>{\PreserveBackslash\raggedleft}p{#1}}
\newcolumntype{L}[1]{>{\PreserveBackslash\raggedright}p{#1}}
\usepackage[accsupp]{axessibility}

\input{preamble}

\definecolor{cvprblue}{rgb}{0.21,0.49,0.74}
\usepackage[pagebackref,breaklinks,colorlinks,citecolor=cvprblue]{hyperref}

\def\paperID{13507} 
\def\confName{CVPR}
\def\confYear{2024}

\title{Learning to Produce Semi-dense Correspondences for Visual Localization}

\author{Khang Truong Giang\textsuperscript{\rm 1} \quad Soohwan Song\textsuperscript{\rm 2}\footnote{Corresponding authors}  \quad Sungho Jo\textsuperscript{\rm 1}\footnotemark[\value{footnote}] \\
\textsuperscript{\rm 1} School of Computing, KAIST, Daejeon, Republic of Korea \\
\textsuperscript{\rm 2} College of AI Convergence, Dongguk University, Seoul, Republic of Korea\\
    {\small*corresponding authors} 
}

\begin{document}
\maketitle
\input{sec/0_abstract}    
\input{sec/1_intro}

\input{sec/2_relatedworks}
\input{sec/3_method}
\input{sec/4_experiments}
\input{sec/5_conclusion}
{
    \small
    \bibliographystyle{ieeenat_fullname}
    \bibliography{main}
}

\input{sec/x_supp}

\end{document}

%% file: preamble.tex
\usepackage[dvipsnames]{xcolor}


%% file: sec/0_abstract.tex
\begin{abstract}
This study addresses the challenge of performing visual localization in demanding conditions such as night-time scenarios, adverse weather, and seasonal changes. While many prior studies have focused on improving image matching performance to facilitate reliable dense keypoint matching between images, existing methods often heavily rely on predefined feature points on a reconstructed 3D model. Consequently, they tend to overlook unobserved keypoints during the matching process. Therefore, dense keypoint matches are not fully exploited, leading to a notable reduction in accuracy, particularly in noisy scenes. To tackle this issue, we propose a novel localization method that extracts reliable semi-dense 2D-3D matching points based on dense keypoint matches. This approach involves regressing semi-dense 2D keypoints into 3D scene coordinates using a point inference network. The network utilizes both geometric and visual cues to effectively infer 3D coordinates for unobserved keypoints from the observed ones. The abundance of matching information significantly enhances the accuracy of camera pose estimation, even in scenarios involving noisy or sparse 3D models. Comprehensive evaluations demonstrate that the proposed method outperforms other methods in challenging scenes and achieves competitive results in large-scale visual localization benchmarks. The code will be available at \href{https://github.com/TruongKhang/DeViLoc}{https://github.com/TruongKhang/DeViLoc}.
\end{abstract}

%% file: sec/1_intro.tex
\section{Introduction}
\label{sec:intro}

\begin{figure*}[t]
  \centering
  \includegraphics[width=0.95\linewidth]{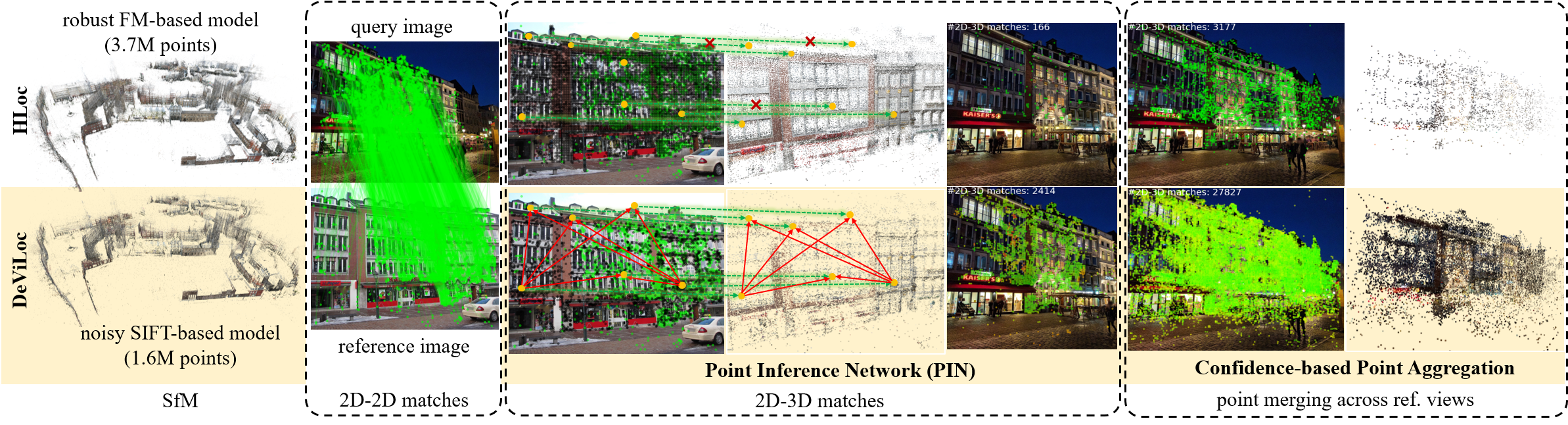}
  \caption{The comparison of the 2D-3D correspondence finding process in our method (DeViLoc) and an existing method (HLoc \cite{sarlin2019coarse}). HLoc heavily relies on a robust 3D point cloud but discards many detected 2D keypoints (depicted in green) during the 2D-3D matching process. In contrast, our method efficiently handles a noisy point cloud through the point inference process of PIN. PIN transforms the entire set of 2D-2D matches into 2D-3D matches. Our method then produces numerous accurate 2D-3D matches across multiple views using a confidence-based aggregation module. These abundant matches significantly enhance localization performance, particularly in scenarios characterized by noisy or sparse 3D point clouds.}
  \label{fig:intro}
\end{figure*}

Visual localization is the process of determining the 6 degrees of freedom (DoF) camera pose for a given query image within a known scene. This fundamental task in computer vision is critical for applications such as robot navigation \cite{lim2012real} and virtual or augmented reality \cite{lynen2015get,middelberg2014scalable}. 
Most leading studies primarily employ a structure-based approach \cite{zeisl2015camera,sattler2016efficient,sarlin2019coarse,brachmann2021visual,sarlin2021back}, consistently exhibiting high localization performance across diverse challenging conditions \cite{svarm2016city,li2012worldwide,sarlin2020superglue,toft2020long}.

Traditionally, structure-based methods heavily rely on feature matching (FM) \cite{schonberger2018semantic,brejcha2020landscapear,sattler2015hyperpoints,taira2018inloc,xue2022efficient}. These methods establish sparse correspondences between 3D points and 2D pixel-level keypoints in images, followed by estimating camera poses using RANSAC-based Perspective-n-Point (PnP). The recent advancements in FM-based methods \cite{sarlin2019coarse,sarlin2020superglue,sun2021loftr} have shown outstanding performance across various benchmarks, particularly in large-scale scenes. However, despite these achievements, FM-based methods encounter substantial challenges in practical scenarios, including dealing with complex lighting conditions, seasonal variations, and changes in perspectives.

Addressing these challenges necessitates a more robust and informative feature-matching approach incorporating detailed 3D points. Current methods, including those achieving semi-dense 2D-2D correspondences through detector-free image matching \cite{sun2021loftr,chen2022aspanformer,giang2023topicfm}, are limited by relying solely on matched sparse features. This limitation persists when considering semi-dense matches, which only account for keypoints observed in a 3D model, neglecting valuable information from unobserved keypoints. Furthermore, FM-based methods demand a detailed 3D point cloud map for accurate localization. Continuously refining 3D feature points for new image inputs is time-intensive, and in some scenarios, localization must be performed with a noisy or sparse 3D point cloud. Overreliance on predetermined 2D and 3D points in the database can lead to a degradation in pose accuracy, especially in noisy cases with texture-less surfaces or repetitive patterns.

Recent studies introduce scene coordinate regression (SCR) methods \cite{brachmann2021visual,brachmann2018learning,Li2019HierarchicalSC,Tang2021LearningCL}, aiming to achieve dense 2D-3D correspondences. Unlike FM-based approaches, SCR methods use an implicit representation of scenes as a learnable function, predicting the dense 3D scene coordinates of a query image. SCR methods excel in compact and stable settings, eliminating the need for storing 3D models; however, they face challenges in dynamic environments and adapting to new viewpoints, limiting their applicability in large-scale scenes. Therefore, there is a need for an alternative method capable of finding dense and accurate 2D-3D correspondences for visual localization.

Therefore, this study proposes a novel FM-based method, semi-\textbf{De}nse \textbf{Vi}sual \textbf{Loc}alization (DeViLoc), aiming to predict dense 2D-3D correspondences for robust and accurate localization. In contrast to existing FM-based methods relying solely on pre-existing 3D points, our method directly converts semi-dense 2D-2D matches into 2D-3D matches. This abundant 2D-3D match information significantly enhances the precision of camera pose estimation, even when dealing with noisy or sparse 3D models. 

This method comprises two main components: 1) the \textit{Point Inference Network} (PIN) and 2) the \textit{Confidence-based Point Aggregation} (CPA) module. PIN plays a crucial role in our method by converting semi-dense 2D-2D matches into 2D-3D matches. It achieves this by directly regressing all 2D keypoints, both observed and unobserved, into 3D scene coordinates. The process involves encoding scene geometry from observed points into latent vectors and propagating 3D information to unobserved positions through attention layers. Next, the CPA module aggregates 2D-3D matches from multiple query-reference pairs, identifying consistent and highly confident 3D points corresponding to the same 2D keypoints in multiple matching views. This step effectively removes outliers from dense matches, and the filtered 2D-3D matches expedite RANSAC-based pose estimation. Ultimately, DeViLoc significantly increases the number of accurate 2D-3D matches for localization.

Fig. \ref{fig:intro} illustrates the 2D-3D matching process of our method and an existing one (HLoc+LoFTR \cite{sarlin2019coarse,sun2021loftr}). The existing method, despite having a dense and precise 3D model, rejects many important points during 2D-3D estimation. In contrast, our method confidently transforms 2D-2D matches into 2D-3D matches, generating numerous matches even in the presence of noisy 3D input and night-time conditions. Consequently, the method yields robust and accurate localization results based on dense matching information, especially in challenging conditions like night-time scenarios, adverse weather, and seasonal changes.

This paper makes the following contributions:
\begin{itemize}
    \item We propose a novel visual localization method that leverages rich matching information by directly converting semi-dense 2D-2D matches into 2D-3D matches. This method significantly improves the accuracy of camera pose estimation, particularly in scenarios with noisy or sparse 3D models.
    \item We introduce a network architecture, Point Inference Network (PIN), designed to directly regress 2D keypoints into 3D points. This network effectively exploits geometric and visual connections between unobserved and observed keypoints, ensuring accurate estimation of 3D information along with associated uncertainties.
    \item We conducted a comprehensive evaluation of our method across diverse datasets. The results indicate that our proposed approach outperforms other state-of-the-art methods in challenging scenes and achieves competitive performance in large-scale visual localization benchmarks. The source code is publicly available.
\end{itemize}

%% file: sec/2_relatedworks.tex
\section{Related Works}
\label{sec:related_works}
Visual localization, which involves estimating camera poses from visual inputs, has been a subject of study for decades \cite{schindler2007city,li2012worldwide,svarm2016city,brachmann2021visual,panek2023visual,kim2023ep2p}. Early approaches \cite{schindler2007city,torii201524,arandjelovic2016netvlad} primarily relied on an image retrieval strategy to directly estimate camera poses from the most similar images in database. This approach is intuitive and efficient, but its performance is significantly influenced by the density of images in the database. To overcome this limitation, an alternative approach learns a neural network to directly predict absolute camera poses from the input images \cite{walch2017image,kendall2015posenet,kendall2017geometric}.

Many studies \cite{zeisl2015camera,svarm2016city,schonberger2018semantic,sattler2016efficient,taira2018inloc,sarlin2019coarse,humenberger2020robust,brachmann2017dsac,brejcha2020landscapear,brachmann2021visual} have shifted their focus towards structure-based approach due to its stability and scalability in diverse scenes. The structure-based methods estimate camera pose by establishing a set of 2D-3D correspondences between image pixels and 3D coordinates of the scene. Leveraging this precise correspondence set, a camera pose can be accurately computed using a PnP solver \cite{gao2003complete,kneip2011novel,persson2018lambda} within a RANSAC paradigm \cite{fischler1981paradigm,barath2018graph,barath2020magsac++}. 
The structure-based methods can be broadly categorized into two main groups: feature matching (FM) and scene coordinate regression (SCR).

\textbf{Feature Matching}. FM-based methods \cite{schonberger2018semantic,brejcha2020landscapear,sarlin2019coarse,li2012worldwide,sattler2015hyperpoints} initially reconstruct a 3D model of the environment from database images using Structure-from-Motion (SfM) \cite{wu2013towards,schonberger2016structure}. Each 3D point in the model is associated with one or several feature descriptors for localization. When a localization request for a query image is made, these methods detect a set of 2D keypoints along with their descriptors and proceed to match them with the 3D points. Several works generate the 2D-3D matches by examining all points in the 3D model \cite{sattler2012improving,li2012worldwide}; however, they face difficulty when dealing with large scenes.

To address this challenge, recent studies \cite{sarlin2019coarse,xue2022efficient,tang2023neumap} have introduced a coarse-to-fine strategy. They initially identify a set of reference images in a database through image retrieval. They then establish 2D-3D matches based on 2D-2D matches between the query and reference images. Subsequent works have focused on enhancing the performance of image matching by incorporating transformers \cite{sarlin2020superglue,sun2021loftr,chen2022aspanformer,giang2023topicfm} or semantic information \cite{xue2023sfd2}. These approaches produce robust and accurate 2D-2D matches, contributing to the robust construction of 3D model and the accurate generation of 2D-3D matches during the localization step. Consequently, FM-based methods have achieved state-of-the-art performance.

However, these methods exhibit inflexibility due to heavily relying on high-fidelity point cloud reconstruction. This time-consuming step is not suitable for online applications such as SLAM or robot navigation, where the 3D point cloud is constructed on-the-fly from a sequence of images. In this situation, the point cloud might be noisy or incomplete, thus degrading localization performance. Furthermore, existing FM-based methods only utilize observed 3D points in the database to generate 2D-3D matches, discarding numerous unrelated 2D-2D matches. This process might compromise the performance in the presence of noisy 3D inputs. On the other hand, our proposed framework aims to adapt to various kinds of 3D inputs and predict semi-dense 2D-3D correspondences.

\textbf{Scene Coordinate Regression}. In contrast to the explicit utilization of 3D models seen in FM-based methods, SCR \cite{brachmann2017dsac,brachmann2021visual,brachmann2018learning,cavallari2019let,cavallari2019real,shotton2013scene,Valentin2016LearningTN} employs an implicit representation of scenes in a form of a machine learning model. This model predicts dense 3D scene coordinates for an input query image. 
While SCR-based methods offer a concise representation of scenes, they encounter challenges when adapting to large-scale scenes, novel scenes, or challenging conditions. Several approaches have been proposed to address these challenges by predicting a scene part-by-part \cite{Brachmann2019ExpertSC,Li2019HierarchicalSC} or employing a coarse-to-fine prediction \cite{Tang2021LearningCL,Yang2019SANetSA}. 
However, these methods still exhibit lower performance compared to FM-based methods. 

Drawing inspiration from SCR, our method aims to predict 3D coordinates for all keypoints in the query image, identified using a detector-free image-matching model. However, in contrast to the dense prediction in SCR, our semi-dense approach reduces the possibility of incorrect 3D prediction by focusing only on regions of interest between the query and reference images. Consequently, our method achieves more accurate performance than SCR-based methods in outdoor scenes.
\begin{figure*}[t!]
  \centering
  \includegraphics[width=0.95\linewidth]{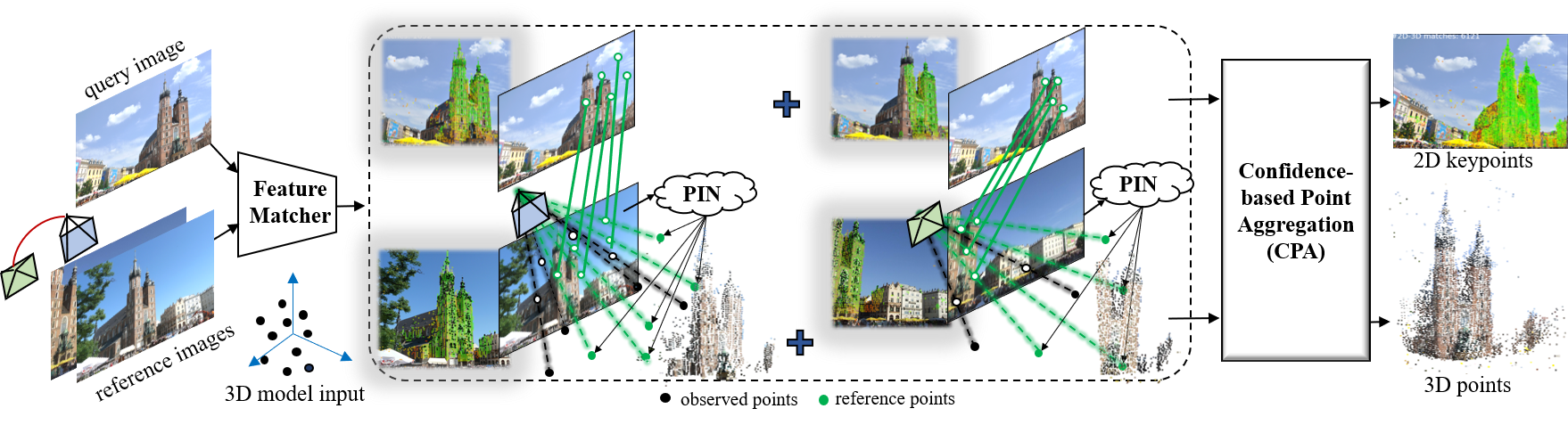}
  \caption{Overview of DeViLoc. First, a feature matcher is employed to detect 2D-2D matches for each pair of query-reference images. Subsequently, the PIN module infers a set of 3D coordinates for all detected 2D keypoints based on the observed data in the reference image. Finally, the CPA module integrates all 2D-3D matches obtained across all query-reference pairs.}
  \label{fig:overview_deviloc}
\end{figure*}

\textbf{3D Scene Representation}. FM-based methods \cite{sarlin2019coarse,sarlin2020superglue,sun2021loftr,giang2023topicfm} have proven to be effective, but they come with a drawback of requiring substantial storage capacity to store both the 3D coordinates and the associated visual features \cite{sarlin2019coarse,sarlin2020superglue}. Consequently, recent studies \cite{Cheng2019CascadedPF,camposeco2019hybrid,yang2022scenesqueezer,panek2022meshloc,brachmann2023accelerated,zhou2022geometry} have shifted their focus toward discovering more space-efficient representations of the scene. 
For example, NeuMap \cite{tang2023neumap} proposed encoding a 3D point cloud into a set of latent codes and then regressing 3D coordinates based on these codes. While these methods successfully reduce storage demands, they discard crucial scene information, thereby limiting their performance. In contrast, this study focuses on improving performance when dealing with challenging point cloud inputs. 

%% file: sec/3_method.tex
\section{Proposed Method}
\label{sec:method}
We follow the coarse-to-fine localization paradigm \cite{sarlin2019coarse} that first retrieves a set of reference images via image retrieval, and then estimates a camera pose by finding 2D-3D correspondences between image pixels and 3D coordinates of the scene. The proposed method aims to address the task of 2D-3D prediction, while applying image retrieval and a PnP pose solver, similar to \cite{sarlin2019coarse}.

\subsection{Overview}
\label{sec:method_overview}
Given a query image $I^q$ and a set of reference images $\{I^{r,1}...I^{r,N_v}\}$ retrieved from a database, our method establishes a set of 2D-3D correspondences $M = \{(k_i,p_i)\}$, where $k_i$ is a 2D keypoint in the query image and $p_i$ represents the corresponding 3D coordinate. Fig. \ref{fig:overview_deviloc} describes the overall architecture of the proposed method, DeViLoc. For each pair of query $I^q$ and reference $I^r$, DeViLoc first extracts local 2D-2D feature correspondences $\{(k_i^q,k_i^r)\}$ using an image-matching network. Here, we employed a detector-free image-matching model \cite{giang2023topicfm} to produce semi-dense correspondences.

The Point Inference Network (PIN) then converts detected 2D keypoints of the reference image into 3D points (Section \ref{sec:method_PIN}). It takes reference image $I^r$ with known camera parameters, along with a sparse set of observed 3D points as inputs, and predicts a set of 3D scene coordinates, $\{p_i\}$, corresponding to all keypoints $\{k_i^r\}$. Based on predicted 3D points $\{p_i\}$ and 2D-2D matches $\{(k_i^q, k_i^r)\}$, we could produce dense 2D-3D matches, $\{(k_i^q,p_i)\}$, for the query image. The basic concept of PIN is inspired by depth completion \cite{wang2023lrru,zhang2023completionformer}, where a dense depth map is reconstructed from a sparse set of observed depth measurements and an input image. However, unlike depth completion, PIN aims to infer a discrete set of depth points, leading to reduced computational costs. Due to this discrete depth prediction, PIN utilizes attention \cite{dosovitskiy2020image} and MLP layers without the need for complex CNNs, as commonly found in conventional depth completion methods. 

Next, the Confidence-based Point Aggregation (CPA) module integrates all 2D-3D matches from multiple query-reference image pairs (Section \ref{sec:method_CPA}). CPA integrates 2D-3D matches with small distances of keypoints into a representative match. It effectively removes outliers by considering the confidence information of each match during the integration process.

\begin{figure*}[t]
  \centering
  \includegraphics[width=0.82\linewidth]{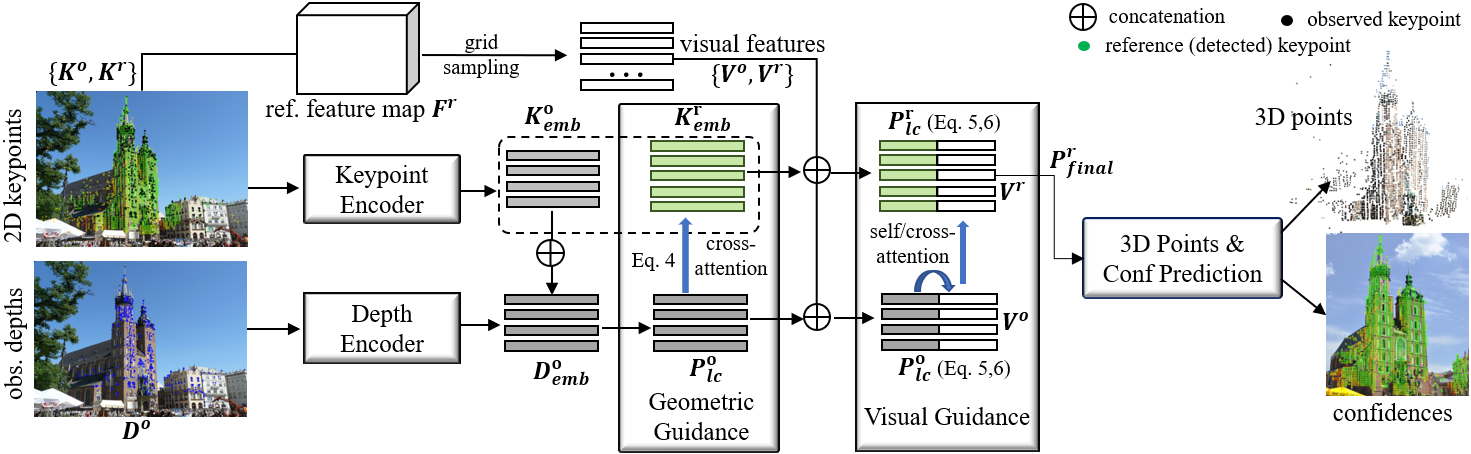}
  \caption{Point Inference Network (PIN). The network begins by learning embeddings for all keypoints $(K_{emb}^o, K_{emb}^r)$ and observed depths $(D_{emb}^o)$. Subsequently, attention layers are employed for both geometric and visual guidance. Finally, the learned latent codes $(P_{lc}^r)$ are utilized to perform regression for the 3D points along with confidence values.}
  \label{fig:pin}
\end{figure*}
\subsection{Point Inference Network}
\label{sec:method_PIN}
For reference image $I^r$, let $O^r=\{o_i^r\}$ be the set of 3D points observed from the constructed 3D model. PIN predicts 3D points $P^r=\{p_i^r\}$ corresponding to detected 2D keypoints $K^r = \{k_i^r\}$ as:
\begin{equation}\label{eq:1}
    P^r = \texttt{PIN}(K^r, O^r, F^r)
\end{equation}
where $K^r \in \mathbb{R}^{N_{points}^r \times 2}$, $O^r \in \mathbb{R}^{N_{points}^o \times 3}$, and $F^r$ is a feature map extracted from image $I^r$ using convolutional networks ($F^r \in \mathbb{R}^{h \times w \times c}$). We directly utilized the output features of the image-matching network \cite{giang2023topicfm} as $F^r$.

The method begins with a preprocessing step where keypoints $K^r$ and observed 3D points $O^r$ are transformed into the same camera-coordinate system using the known camera parameters. Subsequently, we decompose the observed 3D points into observed keypoints $K^o \in \mathbb{R}^{N_{points}^o \times 2}$ and depth values $D^o \in \mathbb{R}^{N_{points}^o \times 1}$. The PIN network then aims to estimate depths $D^r$ for reference keypoints $K^r$, utilizing observed keypoints $K^o$, observed depths $D^o$, and the feature map $F^r$. The network architecture of PIN is illustrated in Fig. \ref{fig:pin}, comprising two fundamental steps to leverage both spatial information and visual similarity for depth estimation: geometric guidance and visual guidance.

To implement geometric guidance, we initially employ two MLP-based encoders to learn embeddings for all 2D keypoint coordinates $\{K^r, K^o\}$, as well as for the observed depths, $D^o$. Additionally, we encode scene geometry from the observed depths by utilizing a self-attention layer.
\begin{align}
    D_{emb}^o &= \texttt{DepthEnc}(D^o) \label{eq:4} \\
    [K_{emb}^r, K_{emb}^o] &= \texttt{KeypointEnc}([K^r, K^o]) \label{eq:5}
\end{align}
The relative position between the observed and unobserved keypoints serves as a primary geometric cue for propagating depth information from observed to unobserved positions. Therefore, we represent the observed positions with latent codes, $P_{lc}^o$, combined from keypoint embeddings $K_{emb}^o$ and depth features $D_{emb}^o$. These latent codes are then passed into a cross-attention layer to generate latent codes $P_{lc}^r$ related to reference keypoints $K_{emb}^r$: 
\begin{equation}
    P_{lc}^r = \texttt{CrsAtt}(K_{emb}^r, P_{lc}^o) \label{eq:7}
\end{equation}
In Eq. \ref{eq:7}, $K_{emb}^r$ is the query, and $P_{lc}^o$ represents the key and value for the cross-attention function, $\texttt{CrsAtt}(.)$.

The geometric guidance mentioned above may lack robustness in estimating accurate depths due to the sparsity of observed data points. Therefore, we also incorporate visual features for more detailed guidance. Considering the feature map $F^r$ in Eq. \ref{eq:1}, we utilize a bilinear grid sampling to extract sets of visual features $V^r$ and $V^o$ for both $K^r$ and $K^o$. These visual features are appended to the latent codes, $P_{lc}^r$ and $P_{lc}^o$. Subsequently, we employ multiple self/cross-attention layers to facilitate visual guidance:
\begin{align}
    P_{lc}^r &= \texttt{Concat}(P_{lc}^r, V^r), P_{lc}^o = \texttt{Concat}(P_{lc}^o, V^o )
    \label{eq:8} \\
    P_{lc}^o &= \texttt{SelfAtt}(P_{lc}^o ), P_{final}^r = \texttt{CrsAtt}(P_{lc}^r, P_{lc}^o) \label{eq:9}
\end{align}
After learning the final latent features, $P_{final}^r$, corresponding to reference keypoints $K^r$, we use two MLP-based networks to predict the depths and confidences:
\begin{subequations}\label{eq:10}
    \begin{equation}
        D^r = \texttt{MLP}(P_{final}^r), \quad C^r = \texttt{MLP}([P_{final}^r, D^r]) \tag{\theequation a,b}
    \end{equation}
\end{subequations}
Finally, the 3D scene coordinates for the reference keypoints are estimated as follows:
\begin{equation}
    P^r=(T^{-1})_{cam \to scene} (D^r [\Hat{K}^r, 1]^T)
\end{equation}
where $T^{-1}$ transforms the 3D points from camera-coordinate to scene-coordinate system. The confidences $C^r$ represent the uncertainty associated with the predicted 3D coordinates, $P^r$. 
\begin{table*}[t]
\centering
\resizebox{.95\textwidth}{!}{%
  \begin{tabular}{|c|m{8em}|ccccccc|ccccc|}
    \hline
    & \multirow{2}{9em}{\textbf{Methods}} & \multicolumn{7}{c|}{\textbf{7scenes (indoor)}} & \multicolumn{5}{c|}{\textbf{Cambridge landmarks (outdoor)}} \\
    \cline{3-14}
    & & Chess & Fire & Heads & Office & Pumpkin & Kitchen & Stairs & Court & King's & Hospital & Shop & St. Mary's \\
    \hline
    \multirow{5}{*}{\rotatebox[origin=c]{90}{D}} & \textcolor{red}{HSCNet} \cite{Li2019HierarchicalSC} & 2/0.7 & 2/0.9 & 1/0.9 & 3/0.8 & \textbf{4/1.0} & 4/1.2 & \textbf{3/0.8} & 28/0.2 & 18/0.3 & 19/0.3 & 6/0.3 & 9/0.3 \\ 
    & \textcolor{red}{DSAC++} \cite{brachmann2018learning} & \textbf{2/0.5} & 2/0.9 & 1/0.8 & \textbf{3/0.7} & \textcolor{cyan}{4/1.1} & 4/1.1 & 9/2.6 & 40/0.2 & 18/0.3 & 20/0.3 & 6/0.3 & 13/0.4 \\ 
    & \textcolor{red}{DSAC*} \cite{brachmann2021visual} & 2/1.10 & 2/1.24 & 1/1.82 & 3/1.15 & 4/1.34 & 4/1.68 & \textcolor{cyan}{3/1.16} & 49/0.3 & 15/0.3 & 21/0.4 & 5/0.3 & 13/0.4 \\
    & SANet \cite{Yang2019SANetSA} & 3/0.88 & 3/1.08 & 2/1.48 & 3/1.00 & 5/1.32 & 4/1.40 & 16/4.59 & 328/1.95 & 32/0.54 & 32/0.53 & 10/0.47 & 16/0.57 \\ 
    & DSM \cite{Tang2021LearningCL}  & \textcolor{cyan}{2/0.68} & \textcolor{cyan}{2/0.80} & 1/0.80 & \textcolor{cyan}{3/0.78} & \textcolor{cyan}{4/1.11} & \textbf{3/1.11} & 4/1.16 & 43/0.19 & 19/0.35 & 23/0.38 & 6/0.30 & 11/0.34 \\
    \hline
    \multirow{5}{*}{\rotatebox[origin=c]{90}{S}} & AS \cite{sattler2016efficient} & 3/0.87 & 2/1.01 & 1/0.82 & 4/1.15 & 7/1.69 & 5/1.72 & \textcolor{cyan}{4/1.01} & 24/0.13 & \textcolor{cyan}{13/0.22} & 20/0.36 & 4/0.21 & 8/0.25 \\ 
    & InLoc \cite{taira2018inloc} & 3/1.05 & 3/1.07 & 2/1.16 & 3/1.05 & 5/1.55 & 4/1.31 & 9/2.47 & - & - & - & - & - \\
    & PixLoc \cite{sarlin2021back} & 2/0.80 & \textbf{2/0.73} & 1/0.82 & 3/0.82 & 4/1.21 & \textcolor{cyan}{3/1.20} & 5/1.30 & 30/0.14 & 14/0.24 & 16/0.32 & 5/0.23 & 10/0.34 \\
    & HLoc[SP+SG] & 2/0.84 & 2/0.93 & \textcolor{cyan}{1/0.74} & 3/0.92 & 5/1.27 & 4/1.40 & 5/1.47 & \textcolor{cyan}{16/0.11} & \textbf{12/0.20} & \textcolor{cyan}{15/0.30} & \textcolor{cyan}{4/0.20} & \textbf{7/0.21} \\
    & HLoc[LoFTR] & 3/0.93 & 2/0.87 & \textcolor{cyan}{1/0.73} & 4/1.02 & 5/1.24 & 5/1.48 & 6/1.47 & 19/0.11 & 16/0.26 & 16/0.29 & \textcolor{cyan}{4/0.21} & 9/0.26 \\ 
    \hline
    \multirow{2}{*}{\rotatebox[origin=c]{90}{SD}} & NeuMap \cite{tang2023neumap} & 2/0.81 & 3/1.11 & 2/1.17 & 3/0.98 & \textcolor{cyan}{4/1.11} & 4/1.33 & 4/1.12 & \textbf{6/0.10} & 14/0.19 & 19/0.36 & 6/0.25 & 17/0.53 \\
    & DeViLoc (Ours) & 2/0.78 & \textbf{2/0.74} & \textbf{1/0.65} & 3/0.82 & \textbf{4/1.02} & \textcolor{cyan}{3/1.19} & 4/1.12 & 18/0.11 & \textbf{12/0.21} & \textbf{13/0.28} & \textbf{4/0.18} & \textcolor{cyan}{7/0.23} \\
    \hline
  \end{tabular}}
  \caption{Evaluation on 7scenes and Cambridge landmarks. The metrics are the median translation (cm) and rotation ($o$) errors. The SCR-based methods highlighted in \textcolor{red}{red} were trained per scene. The best and second-best results are marked in \textbf{bold} and \textcolor{cyan}{cyan}. DeViLoc outperforms the other methods in overall, despite being trained only on MegaDepth.}
  \label{table:7scenes_cambridge}
\end{table*}
\subsection{Confidence-based Point Aggregation}
\label{sec:method_CPA}
PIN produced the 2D-3D correspondences, $(K^q,P) = \{(k_i^q, p_i)\}$, along with their associated confidence values $C^r = \{c_i\}$ for every query-reference pair $(I^q, I^r)$. Subsequently, we aggregated the 2D-3D matches $\{(k_i^q, p_i )^n\}_{n=1…N_v}^{i=1…N_m}$ across $N_v$ sets of matches, with $N_m$ being the number of matches per set. The goal of CPA is to eliminate outliers from the aggregated matches and determine the final matches that exhibit high confidence and consistency. To accomplish this, we started by discarding matches with low confidence through a threshold $\tau$. Next, we grouped adjacent matches using a keypoint quantization step. If the coordinates of two keypoints are closer than $s$ pixels, these keypoints are assigned to the same group. This process is represented by a function $Q_s$, where s denotes the quantization size ($s \in \{2,4\}$ in our experiments). Once matches within the same group were identified, we merged them using a confidence-based averaging operation.

Let $k_j$ be a quantized keypoint, the formulas for the aggregated 3D point $p_j^{agg}$ and the corresponding confidence $c_j^{agg}$ can be written as follows: 
\begin{align}
    p_j^{agg} &= \frac{\sum_{i,n} \textbf{1}(Q_s(k_i^n) = k_j ) c_i^n p_i^n}{\sum_{i,n}\textbf{1}(Q_s(k_i^n) = k_j) c_i^n} \\
    c_j^{agg} &= \frac{\sum_{i,n} \textbf{1}(Q_s(k_i^n ) = k_j) c_i^n}{N_{k_j}}
\end{align}
Here, $\textbf{1}(Q_s(k_i^n) = k_j)$ is a binary indicator, and $N_{k_j} = \sum_{i,n} \textbf{1}(Q_s(k_i^n) = k_j)$ represents the number of keypoints quantized into $k_j$.

\subsection{Loss Functions}
\label{sec:method_loss}
In summary, the proposed approach generates a set of 2D-3D matches, denoted as $M = \{(k_j, p_j^{agg}, c_j^{agg})\}$, for query image $I^q$. Utilizing ground-truth camera matrices $(C^q, T^q)$ and available depth information $D^q = \{d_j^q\}$, we define two functions, incorporating point-matching loss and confidence loss, to train the proposed network.

\textbf{Point-matching loss}. We computed the ground-truth 3D coordinates $\{p_j^{gt}\}$ for the keypoints $\{k_j\}$ using the depths $\{d_j^q\}$ and camera matrices $(C^q,T^q)$. Then, we employed the $L1$ function to calculate the loss between the ground-truth 3D points and predicted 3D points:
\begin{equation}\label{eq:14}
    L_{point}^q = \frac{1}{|M|} \sum_j||p_j^{agg} - p_j^{gt}|| 
\end{equation}

\textbf{Confidence loss}. To train the confidences, we projected the 3D points $\{p_j^{agg}\}$ onto the image plane and assigned a label to each point based on pixel error. If the error between the projected 2D point and the query keypoint $k_j$ is less than $\theta$ pixels ($\theta = 8$ in our experiments), the corresponding confidence $c_j^{agg}$ is labeled as $l_j = 1$, and vice versa. Consequently, the confidence loss, denoted as $L_{conf}^q$, is established through the binary cross-entropy function:
\begin{equation}
    L_{conf}^q = \frac{1}{|M|} \sum_j (l_j \log c_j^{agg} + (1 - l_j) \log (1 - c_j^{agg}))
\end{equation}
The final loss results from the combination of these two loss terms:
\begin{equation}
    L^q = L_{point}^q + \lambda*L_{conf}^q
\end{equation}
Here, $\lambda$ was set to 0.25 in our experiments.

%% file: sec/4_experiments.tex
\begin{table*}[t]
\centering
\resizebox{.99\textwidth}{!}{%
  \begin{tabular}{|c|m{8.8em}|cc|cc|ccc|}
    \hline
    & \multirow{2}{9em}{\textbf{Methods}} & \multicolumn{2}{c|}{\textbf{Aachen Day-Night}} & \multicolumn{2}{c|}{\textbf{RobotCar-Seasons}} & \multicolumn{3}{c|}{\textbf{Extended CMU-Seasons}} \\
    \cline{3-9}
    & & Day & Night & Day-all & Night-all & Urban & Suburban & Park  \\
    \hline
    \rotatebox[origin=c]{90}{D} & ESAC \cite{Brachmann2019ExpertSC} & 42.6 / 59.6 / 75.5 & 6.1 / 10.2 / 18.4 & - & - & - & - & - \\ 
    \hline
    \multirow{10}{*}{\rotatebox[origin=c]{90}{S}} & AS \cite{sattler2016efficient} & 85.3 / 92.2 / 97.9 & 39.8 / 49.0 / 64.3 & 50.9 / 80.2 / 96.6 & 6.9 / 15.6 / 31.7 & 81.0 / 87.3 / 92.4 & 62.6 / 70.9 / 81.0 & 45.5 / 51.6 / 62.0 \\ 
    & D2Net \cite{dusmanu2019d2} & 84.8 / 92.6 / 97.5 & 84.7 / 90.8 / 96.9 & 54.5 / 80.0 / 95.3 & 20.4 / 40.1 / 55.0 & 94.0 / 97.7 / 99.1 & 93.0 / \textcolor{cyan}{95.7} / \textcolor{cyan}{98.3} & \textcolor{cyan}{89.2} / \textcolor{cyan}{93.2} / \textcolor{cyan}{95.0} \\
    & S2DNet \cite{germain2020s2dnet} & 84.5 / 90.3 / 95.3  & 74.5 / 82.7 / 94.9 & 53.9 / 80.6 / 95.8 & 14.5 / 40.2 / 69.7 & - & - & - \\
    & HLoc[SP] \cite{sarlin2019coarse,detone2018superpoint} & 80.5 / 87.4 / 94.2 & 68.4 / 77.6 / 88.8 & 53.1 / 79.1 / 95.5 & 7.2 / 17.4 / 34.4 & 89.5 / 94.2 / 97.9 & 76.5 / 82.7 / 92.7 & 57.4 / 64.4 / 80.4 \\
    & PixLoc \cite{sarlin2021back} & 64.3 / 69.3 / 77.4 & 51.0 / 55.1 / 67.3 & 52.7 / 77.5 / 93.9 & 12.0 / 20.7 / 45.4 & 88.3 / 90.4 / 93.7 & 79.6 / 81.1 / 85.2 & 61.0 / 62.5 / 69.4 \\ 
    & HLoc[SP+SG] & \textbf{89.6} / \textcolor{cyan}{95.4} / \textbf{98.8}  & \textcolor{cyan}{86.7} / \textbf{93.9} / \textbf{100.} & \textbf{56.9} / \textcolor{cyan}{81.7} / \textbf{98.1} & \textcolor{cyan}{33.3} / 65.9 / 88.8 & 95.5 / \textcolor{cyan}{98.6} / \textbf{99.3} & 90.9 / 94.2 / 97.1 & 85.7 / 89.0 / 91.6 \\ 
    & LBR \cite{xue2022efficient} & 88.3 / \textbf{95.6} / \textbf{98.8} & 84.7 / \textbf{93.9} / \textbf{100.} & \textcolor{cyan}{56.7} / \textcolor{cyan}{81.7} / \textbf{98.2} & 24.9 / 62.3 / 86.1 & - & - & - \\
    & HLoc+PixLoc & 84.7 / 94.2 / \textbf{98.8} & 81.6 / \textbf{93.9} / \textbf{100.} & \textbf{56.9} / \textbf{82.0} / \textbf{98.1} & \textbf{34.9} / \textcolor{cyan}{67.7} / \textcolor{cyan}{89.5} & \textbf{96.9} / \textbf{98.9} / \textbf{99.3} & \textcolor{cyan}{93.3} / 95.4 / 97.1 & 87.0 / 89.5 / 91.6 \\
    & HLoc[TopicFM]\cite{giang2023topicfm} & \textcolor{cyan}{88.8} / 94.7 / 97.9 & \textcolor{cyan}{86.7} / 92.9 / \textbf{100.} & - & - & - & - & - \\
    \hline
    \multirow{2}{*}{\rotatebox[origin=c]{90}{SD}} & NeuMap \cite{tang2023neumap} & 80.8 / 90.9 / 95.6 & 48.0 / 67.3 / 87.8 & - & - & - & - & - \\
    & DeViLoc (Ours) & 87.4 / 94.8 / \textcolor{cyan}{98.2} & \textbf{87.8} / \textbf{93.9} / \textbf{100.} & \textbf{56.9} / \textcolor{cyan}{81.8} / \textbf{98.0} & 31.3 / \textbf{68.9} / \textbf{92.4} & \textcolor{cyan}{95.7} / 98.4 / \textbf{99.2} & \textcolor{purple}{\textbf{97.1}} / \textcolor{purple}{\textbf{98.3}} / \textcolor{purple}{\textbf{99.4}} & \textcolor{purple}{\textbf{92.1}} / \textcolor{purple}{\textbf{95.1}} / \textcolor{purple}{\textbf{96.3}} \\
    \hline
  \end{tabular}}
  \caption{Evaluated results on the long-term benchmark \cite{sattler2018benchmarking} using the recall metrics at thresholds of \{$(25cm,2^o)$, $(50cm, 5^o)$, $(5m,10^o)$\}. We compare with various complex baselines that integrate robust FM models into HLoc \cite{sarlin2019coarse} or use PixLoc to refine HLoc’s poses. Our method achieves state-of-the-art performance, especially in the highly challenging localization in the CMU dataset (marked in \textcolor{purple}{\textbf{bold red}}).}
  \label{table:large_scale_benchmark}
\end{table*}
\section{Experiments}
\subsection{Implementation Details}

We conducted experiments on various datasets, including 7scenes \cite{shotton2013scene}, Cambridge \cite{kendall2015posenet}, Aachen Day-Night \cite{sattler2018benchmarking,sattler2012image}, RobotCar Seasons \cite{sattler2018benchmarking,maddern20171}, and Extended CMU Seasons \cite{badino2011visual,toft2020long}. 

\textbf{Detailed pipeline}. To demonstrate the adaptability of the proposed method to noisy and sparse 3D inputs, we utilized 3D point clouds generated through SIFT-based SfM in COLMAP \cite{lowe2004distinctive,schonberger2016structure}. These 3D point clouds are consistently available across all datasets. We employed DenseVLad \cite{torii201524}, NetVLad \cite{arandjelovic2016netvlad}, or CosPlace \cite{berton2022rethinking} to retrieve the top-k reference images. Subsequently, the proposed framework was applied to predict 2D-3D matches from these inputs. During the prediction, we utilized the efficient feature-matching model, TopicFM \cite{giang2023topicfm}, to generate semi-dense 2D-2D matches. Finally, the camera pose was computed from the 2D-3D matches using PnP functions in COLMAP.

\textbf{Training}. Our network was trained on MegaDepth \cite{li2018megadepth}, comprising outdoor scenes from various locations. The trained model was directly used for evaluation across all datasets, eliminating the need for finetuning or retraining. 

\subsection{Evaluation on Cambridge and 7scenes}
We conducted a comparative analysis of our method (\textbf{DeViLoc}) against various state-of-the-art structure-based methods. Based on the characteristics outlined in the related works (Section \ref{sec:related_works}), we primarily organized these methods into three categories: dense (D), sparse (S), and semi-dense (SD) methods. In this classification, SCR-based methods like HSCNet \cite{Li2019HierarchicalSC}, SANet \cite{Yang2019SANetSA}, DSAC* \cite{brachmann2021visual}, DSAC++ \cite{brachmann2018learning}, and DSM \cite{Tang2021LearningCL} were placed in the dense group (D) due to their strategy of making dense 2D-3D predictions. Conversely, methods such as Active Search \cite{sattler2016efficient}, InLoc \cite{taira2018inloc}, and HLoc[SP+SG] \cite{sarlin2019coarse,detone2018superpoint,sarlin2020superglue} executed a sparse matching process between 2D keypoints and the 3D point cloud, classifying them in the sparse group (S). Despite PixLoc \cite{sarlin2021back} not directly detecting 2D keypoints in the query image, its effective utilization of 3D points from a sparse point cloud to find corresponding 2D positions in the query image led us to categorize PixLoc in group S as well. In contrast to both groups (D and S), our method predicts 3D coordinates for all detected keypoints without any point rejection. To the best of our knowledge, NeuMap \cite{tang2023neumap} is the most similar work to ours. Therefore, we assigned DeViLoc and NeuMap to the semi-dense group (SD).

We compared the estimated camera poses of all methods to the ground-truth poses, calculating translation (in cm) and rotation (in degrees) errors \cite{kendall2015posenet} and presenting the median errors for each scene in Table \ref{table:7scenes_cambridge}. The results highlight the effectiveness of dense methods, following the SCR approach, in indoor scenes. This success is attributed to their scene-specific training (e.g., HSCNet, DSAC++, DSAC*) or training on extensive indoor datasets (e.g., DSM, trained on ScanNet \cite{dai2017scannet}). However, these dense methods face challenges in generalizing to outdoor scenes in Cambridge. In contrast, despite being trained on outdoor scenes, our method demonstrates superior performance compared to the dense methods on the 7scenes dataset. The method secures first-place rankings in three scenes (Fire, Heads, Pumpkin) and second-place ranking in one scene (Kitchen). It also outperforms dense methods on the Cambridge dataset.

In comparison to methods in group S, DeViLoc consistently delivers superior performance. Among these, HLoc[SP+SG] is the only method achieving competitive results with our approach on the Cambridge dataset. Notably, HLoc employs a complex pipeline involving re-triangulating SIFT-based point clouds using robust local features and matches detected by SuperPoint (SP) \cite{detone2018superpoint} and SuperGlue (SG) \cite{sarlin2020superglue}. In contrast, our method directly employs the noisy SIFT-based inputs and generally outperforms HLoc on both the 7scenes and Cambridge datasets.  
\subsection{Evaluation on large-scale challenging scenes}
We compared DeViLoc with several contemporary image-matching methods including D2Net \cite{dusmanu2019d2}, SP+SG \cite{detone2018superpoint,sarlin2020superglue}, and TopicFM \cite{giang2023topicfm}, which are frequently incorporated into the HLoc pipeline as robust feature matchers. When using these methods, HLoc requires the extraction of new local features and matches from the database to recalculate the 3D point clouds. Furthermore, we also compared DeViLoc to other methods such as PixLoc \cite{sarlin2021back}, ESAC \cite{Brachmann2019ExpertSC}, and NeuMap \cite{tang2023neumap}, which are not based on the HLoc pipeline. Table \ref{table:large_scale_benchmark} presents the results of all these methods.

\textbf{Aachen}. 
DeViLoc performed comparably with recent FM-based baselines. Additionally, when compared to HLoc[TopicFM], which utilized the same feature matcher, our method exhibited an overall superior performance, demonstrating the effectiveness of the proposed pipeline.

\textbf{RobotCar}. 
As illustrated in Table \ref{table:large_scale_benchmark}, our method surpassed other methods and demonstrated competitive performance compared to HLoc[SP+SG] for the day-time queries. Particularly noteworthy is that DeViLoc significantly outperformed HLoc[SP+SG] for the night-time queries. It is important to highlight that both SP and SG were trained using multiple datasets, while DeViLoc was exclusively trained on MegaDepth \cite{li2018megadepth}. 

\textbf{CMU}. 
Our method significantly outperformed the state-of-the-art pipeline HLoc[SP+SG] with a large margin. Compared to the complex pipeline HLoc+PixLoc, which uses PixLoc to refine the estimated poses of HLoc[SP+SG], DeViLoc improved accuracy by up to 5.1\% on the scene “Park”. This demonstrates the effectiveness and stability of our approach in difficult localization conditions.

\subsection{Ablation Study}
\begin{figure}[t]
  \centering
  \includegraphics[width=0.9\columnwidth]{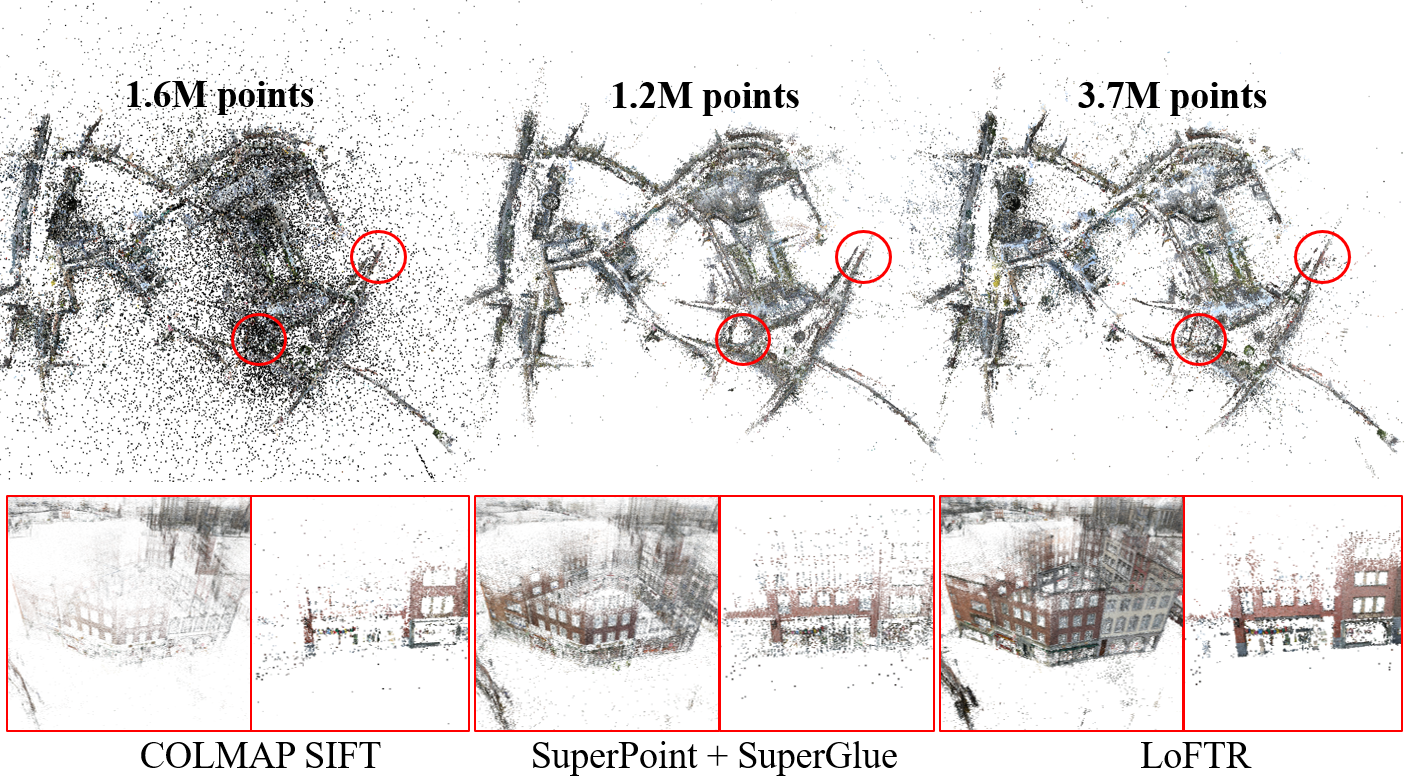}
  \caption{Comparison between point clouds built from traditional FM (SIFT \cite{lowe2004distinctive}), sparse FM (SP+SG \cite{detone2018superpoint,sarlin2020superglue}), and detector-free FM (LoFTR \cite{sun2021loftr}). DeViLoc can handle well the noisy SIFT-based input to achieve competitive performance compared to the precise (SP+SG) or dense (LoFTR) inputs (shown in Table \ref{table:ablation_pointclouds}).}
  \label{fig:ablation_pointclouds}
\end{figure}

\textbf{Evaluating the performance with noisy and sparse inputs}. To assess how effectively DeViLoc handles noisy and sparse inputs, we conducted an experiment using the Aachen Day-Night dataset. Our pipeline utilized 3D point cloud inputs generated by different image-matching models, including SIFT, SP+SG, and LoFTR. As illustrated in Fig. \ref{fig:ablation_pointclouds}, the 3D point cloud generated by SIFT exhibits significantly more noise and sparsity compared to those produced by the SP+SG and LoFTR models. The results for each input model are presented in Table \ref{table:ablation_pointclouds}. We observe that the precise and dense 3D inputs from SP+SG or LoFTR only slightly improve performance. This highlights the adaptability of our approach in handling various types of 3D inputs, proving effective even in the presence of noisy and sparse data.

\begin{figure}[t]
  \centering
  \includegraphics[width=0.9\linewidth]{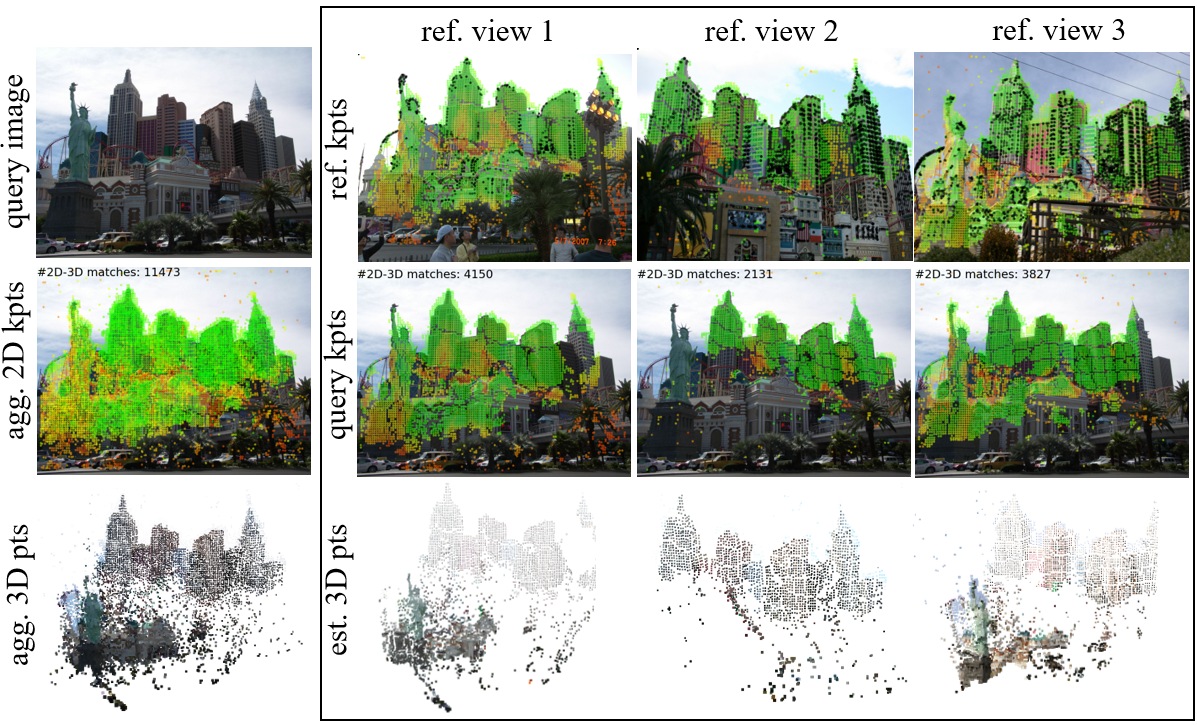}
  \caption{Illustration of 2D-3D correspondences estimated by DeViLoc for several pairs of images. The observed 2D keypoints are marked in black, while the reference keypoints are represented in orange (low confidence) or green (high confidence).} 
  \label{fig:ablation_viz}
\end{figure}

\textbf{Visualization of semi-dense matching}. Fig. \ref{fig:ablation_viz} illustrates the detected 2D keypoints and the corresponding 3D points produced by DeViLoc. We visualized multiple pairs of query and reference views. 
Despite a higher number of detected keypoints compared to the observed keypoints (especially in reference view 1), our method is capable of effectively estimating 3D points along with their uncertainties. Notably, points in the sky or near-edge regions of the scene tend to exhibit lower confidence. Ultimately, our method yields a significant number of 2D-3D matches after the point aggregation step, substantially improving performance, particularly in challenging scenes.
\begin{table}[t]
\centering
\resizebox{.83\columnwidth}{!}{%
  \begin{tabular}{|m{9em}|cc|}
    \hline
    \multirow{2}{*}{\textbf{Models}} & \textbf{Day} & \textbf{Night} \\
    \cline{2-3}
    & \multicolumn{2}{c|}{(0.25m,$2^o$) / (0.5m,$5^o$) / (5.0m,$10^o$)} \\
    \hline
    DeViLoc[half-SIFT] & \textcolor{cyan}{87.5} / 94.1 / 97.9 & 86.7 / \textcolor{cyan}{92.9} / \textbf{100.} \\
    DeViLoc[SIFT] & 87.4 / \textcolor{cyan}{94.8} / \textcolor{cyan}{98.2} &  \textcolor{cyan}{87.8} / \textbf{93.9} / \textbf{100.} \\
    DeViLoc[SP+SG] & 87.3 / \textbf{95.3} / \textbf{98.3} & \textbf{88.8} / \textcolor{cyan}{92.9} / \textbf{100.} \\
    DeViLoc[LoFTR] & \textbf{87.9} / \textcolor{black}{94.7} / \textcolor{cyan}{98.2} & \textbf{88.8} / \textcolor{cyan}{92.9} / \textbf{100.} \\
    \hline
  \end{tabular}}
  \caption{Ablation study of DeViLoc on Aachen Day-Night when using different point cloud inputs.}
  \label{table:ablation_pointclouds}
\end{table}

\textbf{Effectiveness of proposed modules}. 
We implemented three models to measure the contributions of the proposed modules, PIN (Section \ref{sec:method_PIN}) and CPA (Section \ref{sec:method_CPA}):
\begin{itemize}
    \item Model-A: This model executes the standard 2D-3D matching process without utilizing the PIN and CPA modules, akin to existing pipelines like HLoc \cite{sarlin2019coarse}.
    \item Model-B: This model only integrates the PIN module, generating semi-dense matches. All these matches are directly fed into the PnP solver for pose estimation.
    \item Model-C: This model combines both PIN and CPA, including the entire process of the proposed method.
\end{itemize}
Table \ref{table:ablation_modules} presents the results of pose estimation using AUC with thresholds of $\{2^o, 5^o, 10^o\}$ \cite{sarlin2020superglue}. The findings indicate that the proposed modules (Model-B and Model-C) significantly enhance performance as compared to the baseline (Model-A). Notably, these modules do not impose a substantial runtime burden, requiring only about $100ms$ to generate a more extensive set of 2D-3D matches, ultimately enhancing overall performance.

\textbf{Impact of the top-k reference images}. Table \ref{table:ablation_modules} shows that employing more reference images can enhance performance. However, this improvement comes at the expense of significantly increased runtime and the number of matches.

\textbf{Using different feature matcher}. We evaluated the performance by substituting the detector-free TopicFM \cite{giang2023topicfm} with the detector-based SP+SG. The AUC metrics did not decrease significantly, as indicated in Table \ref{table:ablation_modules}.
\begin{table}[t]
\centering
\resizebox{.99\columnwidth}{!}{%
  \begin{tabular}{|l|c|c|c|c|}
    \hline
     \multirow{2}{*}{\textbf{Model}} & \textbf{AUC} & \textbf{Time} & \textbf{\#P} & \multirow{2}{*}{\textbf{\#M}} \\
     & ($2^o / 5^o / 10^o$) & (s) & $\times 10^6$ & \\
    \hline
    A[SIFT] (A $\Leftrightarrow$ HLoc) & 72.4 / 85.5 / 91.1 & 0.49 & 11.6 & 42 \\ 
    A[SP+SG] & 72.8 / 86.4 / 92.2 & 0.53 & 11.6 & 54 \\
    A[LoFTR] & 74.7 / 87.7 / 93.2 & 0.51 & 11.6 & 297 \\
    B (A+PIN) & 78.8 / 90.0/ 94.7 & 0.56 & 15.9 & 4743 \\
    C (A+PIN+CPA)$\Leftrightarrow$DeViLoc & \textbf{79.6 / 90.5 / 94.9} & 0.57 & 15.9 & 4246 \\
    \hline
    C (TopicFM$\rightarrow$SP+SG) & 78.2 / 89.9 / 94.7 & 1.02 & 17.6 & 993 \\
    \hline
    DeViLoc (top-5) & 81.0 / 91.4 / 95.5 & 0.90 & 15.9 & 7668 \\
    DeViLoc (top-10) & \textbf{82.6 / 92.2 / 96.0} & 1.92 & 15.9 & 13970 \\
    \hline
  \end{tabular}}
  \caption{Effectiveness of the proposed PIN and CPA (top), impact of the feature matcher (middle), and ablation of the top-k image retrieval (bottom) on MegaDepth \cite{li2018megadepth}. The top-3 retrieval is used by default. Except for Model-A, the others were tested with SIFT inputs. \textbf{P} is the model parameters and \textbf{M} is the 2D-3D matches.}
  \label{table:ablation_modules}
\end{table}

%% file: sec/5_conclusion.tex
\section{Conclusions and Limitations}
This study introduces a robust structure-based framework for visual localization that minimizes reliance on the precise reconstruction of 3D point clouds. Our approach exhibits stable performance even when confronted with sparse and noisy 3D inputs. To achieve this, we present two novel modules: the Point Inference Network and the Confidence-based Point Aggregation. Consequently, the method generates numerous 2D-3D correspondences, leading to significant enhancements in challenging conditions, including textureless scenes, large-scale environments, and variations in weather and seasons. However, the computational efficiency of our proposed method has limitations. The runtime experiences a slowdown as the number of matching pairs between query and reference images increases. Addressing this limitation will be a focus of our future work.
\section{Acknowledgments}
This work was supported by the Ministry of Trade, Industry and Energy (MOTIE, Korea) under Grant 20007058, and by the National Research Foundation of Korea (NRF) funded by the Korean Government (MSIT) under Grant RS-2023-00208052.

%% file: sec/x_supp.tex
\clearpage
\maketitlesupplementary
\section{Method Details}
\subsection{Attention Layers}
We implemented a standard transformer block \cite{dosovitskiy2020image} for the self-/cross-attention layer. However, we omitted the conventional positional encoding step because our proposed network explicitly learned keypoint embeddings to guide depth estimation. We utilized one cross-attention layer and three pairs of self and cross-attention layers to perform geometric (Eq. 7) and visual guidance (Eq. 9), respectively.
\subsection{Normalization Techniques}
During our experiments, we observed that applying normalization techniques to depth and loss enhances the training network's convergence speed and increases overall generalizability. Specifically, we extracted the minimum and maximum depth values, $d_{min}$ and $d_{max}$, from the observed depth $D^o$. We normalized $D^o$ before applying Eq. 4. Subsequently, we converted the output depth $D^r$ in Eq. 10a back into the original depth range using $d_{min}$ and $d_{max}$.

To normalize the loss in Eq. 14, we computed the standard deviation ($\sigma$) of ground-truth depths and used it as a scaling factor.
\subsection{Implementation of CPA}
While the PIN module (Section 3.2) is naturally differentiable when using standard neural network layers in Pytorch, it is required to make the CPA module (Section 3.3) differentiable with respect to the predicted 3D points and confidences of PIN. This ensures that our network can be trained end-to-end across multiple reference views.

The quantization function $Q_s$, which approximates keypoint $k_i$ within cell size $s$, is simply designed as $Q_s(k_i) = \texttt{round}(\frac{k_i}{s}) * s$, where $\texttt{round}(x)$ converts $x$ into the closest integer number. 

To efficiently implement the point aggregation step in Eqs. 12\&13, we extracted a set of unique quantized keypoints and then used the $\texttt{index\_reduce}$ function in Pytorch, as follows:
{\footnotesize\begin{verbatim}
 uniq_kpts2d, indices = torch.unique(Q_s(kpts2d))
# aggregate confidence values (Eq. 13)
 agg_conf.index_reduce_(0, indices, conf, "mean")
# aggregate 2D keypoints
 agg_kpts2d.index_reduce_(0, indices, \
                          kpts2d*conf, "mean")
 agg_kpts2d = agg_kpts2d / (agg_conf + 1e-6)
# aggregate 3D points (Eq. 12)
 agg_pts3d.index_reduce_(0, indices, \
                         pts3d*conf, "mean")
 agg_pts3d = agg_pts3d / (agg_conf + 1e-6)
\end{verbatim}}
\begin{figure*}[h]
  \centering
  \includegraphics[width=0.85\linewidth]{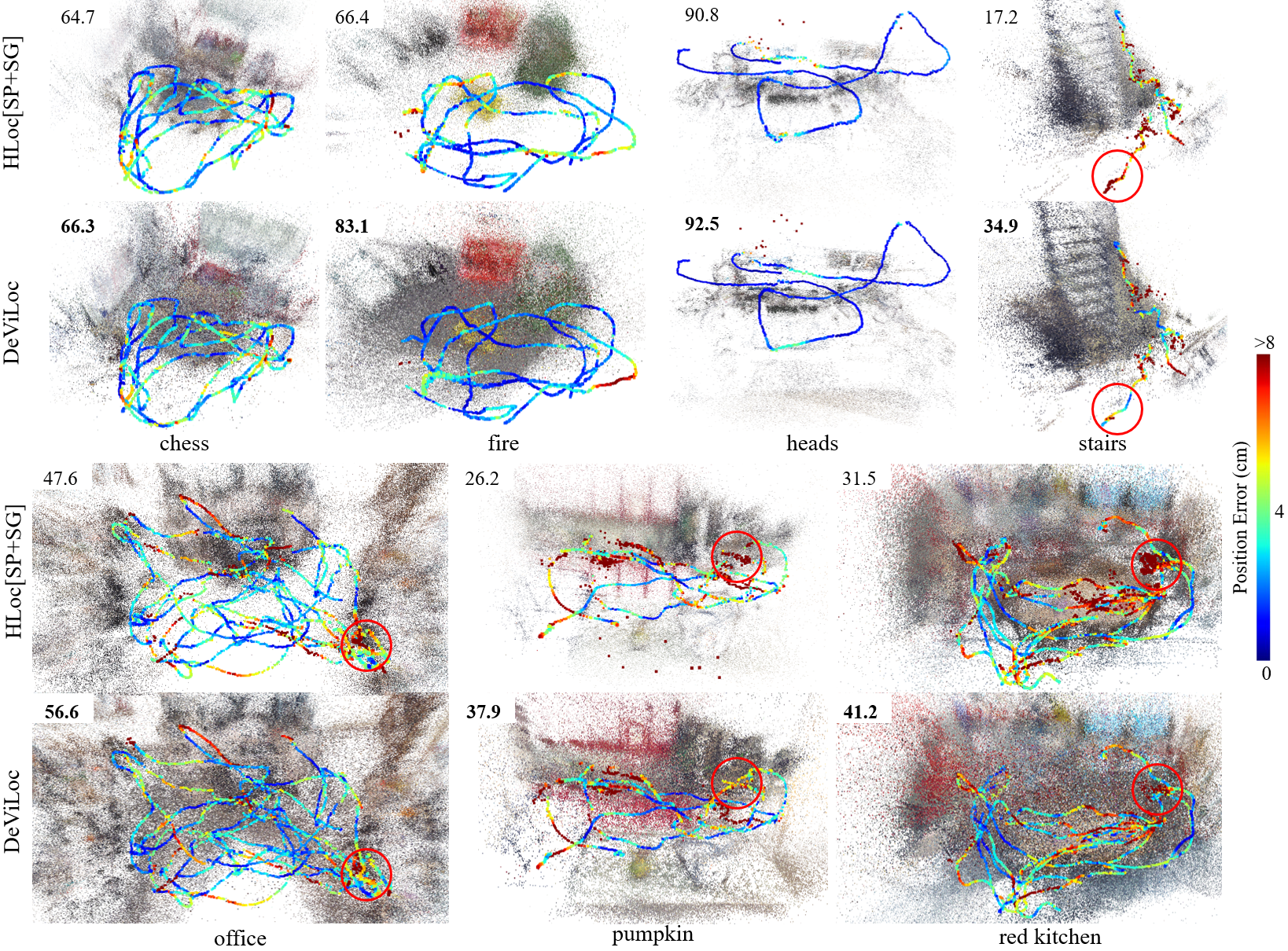}
  \caption{Qualitative results on 7scenes. We visualize the estimated camera positions for all query images with highlighted position errors using a color map. The point cloud inputs for each method are displayed in the background. Additionally, we provide the percentage of query images with a camera pose error below (3cm, $3^o$). Despite having to deal with very noisy 3D inputs, DeViLoc consistently outperforms HLoc[SP+SG] across most scenes.}
  \label{fig:supp_7scenes}
\end{figure*}
\begin{figure*}[h]
  \centering
  \includegraphics[width=0.99\linewidth]{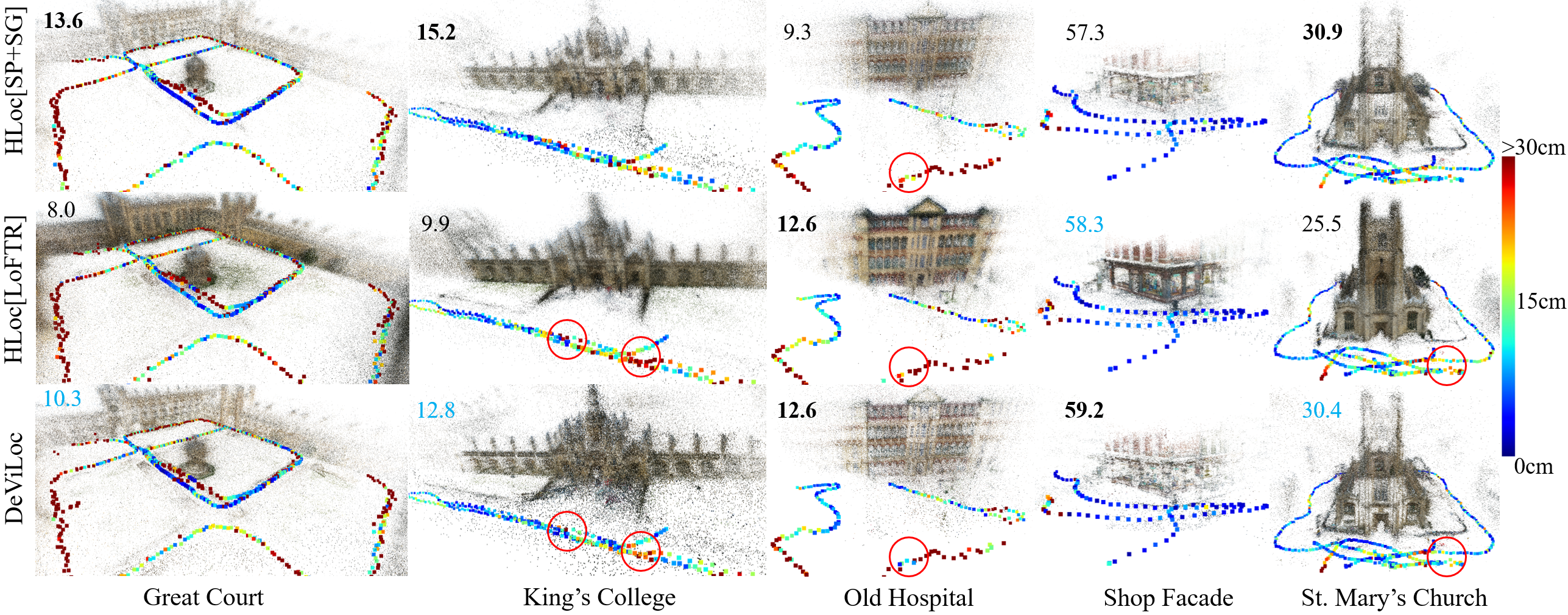}
  \caption{Qualitative evaluation for the Cambridge dataset. We compare with robust FM-based methods, HLoc[SP+SG] \cite{detone2018superpoint,sarlin2020superglue} and HLoc[LoFTR] \cite{sun2021loftr}. The position errors are color-coded, and the percentage of queries with the pose threshold of ($5$cm, $5^o$) is reported. DeViLoc outperforms HLoc[LoFTR] and achieves competitive performance compared to HLoc[SP+SG], even when handling more sparse and noisy inputs.}
  \label{fig:supp_cambridge}
\end{figure*}
\section{Detailed Experiments}
\subsection{Training} 
The proposed method was developed using PyTorch and optimized with AdamW. We conducted training on MegaDepth \cite{li2018megadepth}, comprising 196 scenes from various global locations. For each scene, we randomly selected the top 300 query images and the top 3 reference images for each query. During evaluation, the number of reference images can be increased to enhance performance. We directly utilized pre-trained weights for the 2D feature-matching model and updated weights exclusively for the proposed PIN and CPA (Sections 3.2\&3.3). The training process utilized two NVIDIA GeForce RTX 3090 GPUs, each equipped with 24GB of memory. The initial learning rate and batch size were set to 0.001 and 4, respectively. The training spanned 40 epochs and was completed within two days. The model, trained on MegaDepth, was directly used for evaluation across all datasets, eliminating the need for finetuning or retraining.

\subsection{Analysis for 7Scenes and Cambridge}
\quad \textbf{7scenes}. The 7scenes dataset, primarily designed for indoor scenarios, introduces several challenges, including textureless surfaces and repetitive patterns. Therefore, the SfM pipeline usually produces noisy 3D models. As illustrated in Fig. \ref{fig:supp_7scenes}, the generated point clouds for multiple scenes exhibit substantial noise, failing to accurately represent the scene structures. In our evaluations of the dataset, we employed an image resolution of $640 \times 480$ and configured the quantization size ($s$) in the CPA module to 2. Similar to the setups of HLoc \cite{sarlin2019coarse,sarlin2020superglue}, we employed DenseVLad \cite{torii201524} to select the top 10 reference images for each query image.

The quantitative results using median errors were reported in Table 1 of the main paper. Here, we further provide a qualitative evaluation in Fig. \ref{fig:supp_7scenes}. We visualized the point cloud inputs and drew trajectories of the estimated camera positions. As shown in Fig. \ref{fig:supp_7scenes}, the point clouds of DeViLoc contain a considerable amount of noisy points compared to HLoc. However, DeViLoc can predict more accurate camera positions, as depicted by the color codes. We also calculated the percentage of test images in which their camera positions and orientations are smaller than $3$cm and $3^o$ respectively. The results indicate that our method is significantly better than the robust pipeline HLoc[SP+SG].

\begin{table*}[t]
\centering
\resizebox{.99\textwidth}{!}{%
  \begin{tabular}{m{6em}|C{5.2em}C{5.2em}C{5.2em}C{5.2em}C{5.2em}C{5.2em}C{5.2em}C{5.2em}C{5.2em}C{5.2em}C{5.2em}}
    \hline
    \textbf{Method} & urban & suburban & park &  overcast & sunny & foliage & mixed fol. & no foliage & low sun & cloudy & snow \\
    \hline
    AS \cite{sattler2016efficient} &81.0/87.3/92.4 & 62.6/70.9/81.0 & 45.5/51.6/62.0 & 64.1/70.8/78.6 & 55.2/62.3/71.3 & 58.8/65.3/73.9 & 59.2/67.5/77.4 & 83.3/88.9/94.6 & 65.8/73.4/82.8 & 71.6/77.6/84.2 & 73.0/81.0/90.5 \\ 
    D2Net \cite{dusmanu2019d2} &94.0/97.7/99.1	& 93.0/\textcolor{cyan}{95.7}/\textcolor{cyan}{98.3} & \textcolor{cyan}{89.2}/\textcolor{cyan}{93.2}/\textcolor{cyan}{95.0} & 92.5/\textcolor{cyan}{95.6}/\textcolor{cyan}{97.5} & 86.2/91.8/\textcolor{cyan}{95.2} & 88.0/92.8/\textcolor{cyan}{95.9} & \textcolor{cyan}{94.3}/\textcolor{cyan}{96.9}/\textcolor{cyan}{98.4} & \textcolor{cyan}{98.0}/\textcolor{cyan}{99.4}/\textbf{99.8} & \textcolor{cyan}{95.1}/\textcolor{cyan}{97.5}/\textcolor{cyan}{98.7} & 95.5/\textcolor{cyan}{97.5}/\textcolor{cyan}{98.9} & \textcolor{cyan}{97.2}/\textcolor{cyan}{98.9}/\textbf{99.6} \\
    HLoc[SP] \cite{detone2018superpoint} &  89.5/94.2/97.9 & 76.5/82.7/92.7 & 57.4/64.4/80.4 & 77.1/82.8/91.8 & 65.1/72.3/86.8 & 69.2/75.5/88.3 & 75.2/81.7/90.8 & 88.7/92.8/96.4 & 78.0/83.9/91.8 & 83.4/87.7/94.0 & 80.7/86.6/93.2 \\
    R2D2 \cite{NEURIPS2019_3198dfd0} & 89.7/96.6/98.3 & 76.1/83.8/89.0 & 64.4/72.1/76.5 & 79.9/87.0/90.6 & 70.3/78.3/83.2 & 74.1/81.2/85.6 & 75.7/84.1/87.9 & 86.6/93.3/95.3 & 77.8/85.7/89.3 & 84.1/90.0/92.5 & 79.8/87.6/91.1 \\
    PixLoc \cite{sarlin2021back} & 88.3/90.4/93.7 & 79.6/81.1/85.2 & 61.0/62.5/69.4	& 78.5/80.0/84.8 & 72.4/75.6/81.8 & 74.3/76.8/82.7 & 76.6/77.7/81.8	& 84.1/84.8/87.7 & 77.5/78.4/82.2	& 82.0/82.9/87.0 & 75.7/76.4/80.4 \\ 
    HLoc[SP+SG] &  95.5/\textcolor{cyan}{98.6}/\textbf{99.3} & 90.9/94.2/97.1 & 85.7/89.0/91.6 & 92.3/95.3/96.9 & 86.1/91.3/94.6 & 88.3/92.5/95.3 & 91.6/94.5/96.2 & 95.4/97.1/98.3 & 91.8/94.4/96.3 & 95.2/97.0/\textcolor{cyan}{98.0} & 92.3/94.6/96.6 \\
    HLoc+PixLoc & \textbf{96.9/98.9/99.3} & \textcolor{cyan}{93.3}/95.4/97.1 & 87.0/89.5/91.6 & \textcolor{cyan}{93.6}/95.5/96.9 & \textcolor{cyan}{88.4}/\textcolor{cyan}{92.4}/94.6 & \textcolor{cyan}{90.3}/\textcolor{cyan}{93.4}/95.3 & 93.3/95.0/96.2 & 96.6/97.5/98.3 & 93.6/95.1/96.3 & \textcolor{cyan}{96.2}/97.3/\textcolor{cyan}{98.0} & 94.3/95.3/96.6 \\
    SFD2 \cite{xue2023sfd2} & 95.0/97.5/98.6 & 90.5/92.7/95.3 & 86.4/89.1/91.2 & 92.1/94.0/95.8 & 86.3/90.3/93.4 & 87.9/91.0/93.9 & 91.9/94.0/95.5 & 95.3/96.6/97.6 & 92.4/94.4/95.8 & 93.3/94.7/96.3 & 92.9/94.6/96.0 \\
    \hline
    Ours (top-10) & \textcolor{cyan}{95.7}/98.4/\textcolor{cyan}{99.2} & \textbf{97.1/98.3/99.4} & \textbf{92.1/95.1/96.3} & \textbf{96.2/97.9/98.7} & \textbf{90.4/94.8/96.8}	& \textbf{92.2/95.5/97.2} & \textbf{97.1/98.6/99.2} & \textbf{98.5}/\textcolor{cyan}{99.3}/\textcolor{cyan}{99.6} & \textbf{97.4/98.7/99.2} & \textbf{97.4/98.3/98.9} & \textbf{97.7}/\textcolor{cyan}{98.6}/\textcolor{cyan}{99.1} \\
    Ours (top-15) & \textcolor{cyan}{96.5}/\textbf{98.9}/\textbf{99.3} & \textbf{97.7}/\textbf{98.8}/\textbf{99.8} & \textbf{93.1}/\textbf{96.1}/\textbf{97.1} & \textbf{96.7}/\textbf{98.2}/\textbf{98.9} & \textbf{91.9}/\textbf{96.3}/\textbf{98.0}	& \textbf{93.4}/\textbf{96.7}/\textbf{98.2} & \textbf{97.5}/\textbf{98.8}/\textbf{99.2} & \textbf{98.9}/\textbf{99.6}/\textcolor{cyan}{99.7} & \textbf{97.8}/\textbf{98.9}/\textbf{99.2} & \textbf{98.0}/\textbf{98.9}/\textbf{99.3} & \textbf{98.4}/\textbf{99.1}/\textbf{99.4} \\
    \hline
  \end{tabular}}
  \caption{Detailed breakdown of the results on the extended CMU dataset. The results are categorized into different scenarios such as scene types (urban, suburban, park), weather conditions (overcast, sunny, low sun, cloudy, snow), and foliage appearance (foliage, no foliage, mixed foliage). The metrics are calculated with pose error thresholds of \{$(25\text{cm},2^o)$, $(50\text{cm}, 5^o)$, $(5\text{m},10^o)$\}. We additionally include the other FM-based methods, R2D2 \cite{NEURIPS2019_3198dfd0} and SFD2 \cite{xue2023sfd2}, for comparison. As depicted in the table, DeViLoc demonstrates state-of-the-art performance.}
  \label{table:supp_cmu}
\end{table*}
\textbf{Cambridge landmarks}. The Cambridge dataset includes five outdoor scenes, featuring query and reference images taken from various trajectories. We utilized NetVLad \cite{arandjelovic2016netvlad} to retrieve 20 reference images for each query image. In the testing phase, the longest dimension of an image was resized to 864, and the quantization size (s) was configured to 4. For both datasets, a RANSAC threshold of 20 pixels was employed in the PnP solver during the camera pose estimation step. 

The qualitative results on Cambridge are shown in Fig. \ref{fig:supp_cambridge}. We compared with two FM-based methods, HLoc[SP+SG] and HLoc[LoFTR \cite{sun2021loftr}]. We found that although the detector-free method LoFTR can produce dense point cloud inputs, its performance is not significantly improved in comparison to the detector-based method, SP+SG. The main reason is that the dense point clouds produced by HLoc[LoFTR] also increase the number of imprecise 3D points. Meanwhile, the transformer-based model, SuperGlue (SG), is very effective in eliminating noisy points. In contrast to both models, our point clouds built from SIFT-based COLMAP are both noisy and sparse. However, our method is still better than HLoc[LoFTR] and achieved competitive performance compared to HLoc[SP+SG].

\subsection{More Details of Large-Scale Evaluation}
We conducted additional evaluations of DeViLoc on an extensive visual localization benchmark that includes long-term and large-scale datasets \cite{sattler2018benchmarking}. Our evaluations were performed on three datasets, Aachen Day-Night \cite{sattler2018benchmarking,sattler2012image}, RobotCar-Seasons \cite{sattler2018benchmarking,maddern20171}, and Extended CMU-Seasons \cite{toft2020long,badino2011visual}. The predicted camera poses were submitted to the benchmark website (\href{https://www.visuallocalization.net}{https://www.visuallocalization.net}) to obtain recall metrics at thresholds of $(25\text{cm},2^o)$,$(50\text{cm}, 5^o)$, and $(5\text{m},10^o)$.

\textbf{Aachen}.  The Aachen Day-Night dataset includes 4328 database images from various locations in Aachen city, along with 922 query images captured under both day and night conditions. In this evaluation, we selected the top-50 reference images per query using NetVLad \cite{arandjelovic2016netvlad}. We resized the longest dimension of each image to 864 and applied the quantization size of 4. We then used a RANSAC threshold of 15 pixels to estimate camera poses. 

Although our method experienced a slightly inferior performance compared to HLoc[SP+SG] as shown in Table 2 of the main paper, it still demonstrated the effectiveness in in handling low-quality point cloud input and challenging night-time conditions.
\begin{table}[t]
\centering
\resizebox{.95\columnwidth}{!}{%
  \begin{tabular}{|c|c|c|}
    \hline
      & Day-all & Night-all \\
    \hline
    DeViLoc w/o CPA $(\tau=0)$ & \textbf{57.0} / 81.7 / 97.4 & 27.1 / 67.3 / \textbf{92.8} \\ 
    DeViLoc w/ CPA $(\tau=0.5)$ & \textcolor{cyan}{56.9} / \textcolor{cyan}{81.8} / \textbf{98.0} & \textcolor{cyan}{31.3} / \textcolor{cyan}{68.9} / 92.4 \\
    DeViLoc w/ CPA $(\tau=0.8)$ & 56.2 / \textbf{82.0} / \textbf{98.0} & \textcolor{purple}{\textbf{32.5}} / \textcolor{purple}{\textbf{69.8}} / \textcolor{cyan}{92.5} \\
    \hline
  \end{tabular}}
  \caption{Effectiveness of CPA module on the RobotCar dataset.}
  \label{table:updated_robotcar}
\end{table}

\textbf{RobotCar}. The RobotCar-Seasons dataset is notably challenging, featuring 26121 database images and 11934 query images taken in different seasons (winter, summer), various weather conditions (sun, rain, snow), and different times of the day (dusk, dawn, night). We employed the top-20 reference images, maintaining similar settings for image size, quantization size, and RANSAC threshold as in the Aachen evaluation. 

\begin{figure*}[h]
  \centering
  \includegraphics[width=0.81\linewidth]{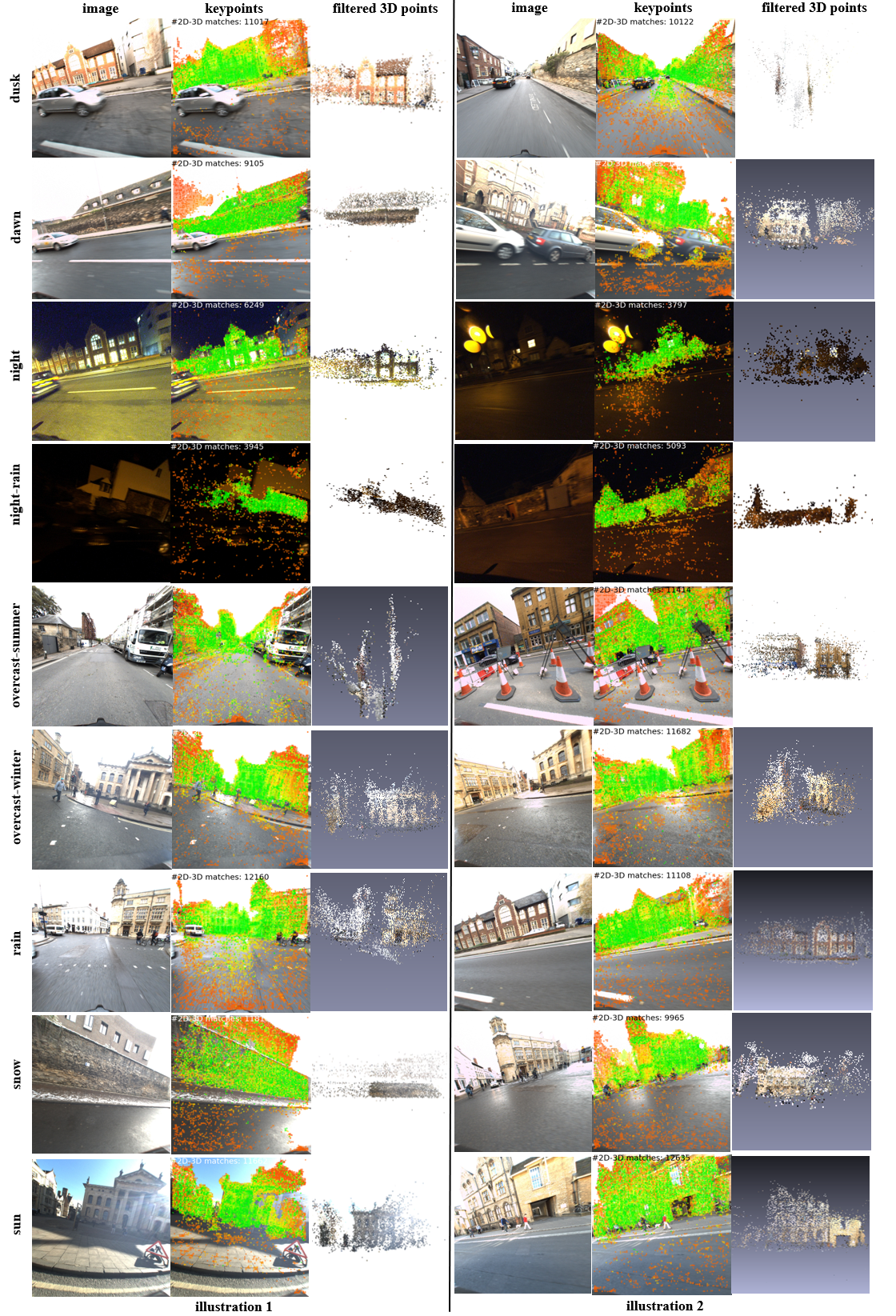}
  \caption{Illustrations of our 2D-3D matching results on the RobotCar dataset under challenging conditions, including varying times (dusk, dawn, night), weather conditions (rain, snow, sun), and seasons (summer, winter). DeViLoc successfully predicts numerous matches with uncertainties, effectively filtering out low-confidence matches (depicted in orange) and enhancing overall matching results.}
  \label{fig:supp_robotcar}
\end{figure*}

We showcased the effectiveness of our method in producing semi-dense 2D-3D matches and filtering noise on the RobotCar dataset. Fig. \ref{fig:supp_robotcar} presents the 2D-3D matching results under various localization conditions. We observed that our method demonstrates proficiency in estimating highly confident matches, particularly in areas associated with buildings. This capability stems from our model being trained on the MegaDepth dataset. Leveraging the confidence estimation module, our approach adeptly eliminates inaccurate matches, as depicted in the orange colors. Additionally, we conducted a quantitative experiment with different setups of confidence threshold $\tau$. Table \ref{table:updated_robotcar} demonstrates the effectiveness of noise filtering using the CPA module.

However, there exist several cases in which our CPA module can not successfully eliminate noisy 2D-3D matches. For instance, as depicted in Fig. \ref{fig:supp_robotcar}, some failure cases include \textit{(overcast-winter, illustration 1\&2)} and \textit{(rain, illustration 1)}, characterized by depth predictions at extended distances. This led to a subpar performance of our method on the RobotCar dataset compared to HLoc in one metric (Table 2).

\textbf{CMU}. The extended CMU dataset provides a more challenging localization scenario with a high number of query images (56613) captured from different locations and conditions. The dataset contains 60937 database images. Building a point cloud input from these images requires days to finish. For FM-based methods that utilize the HLoc pipeline, it is required to build the point clouds based on the detected features. Our method did not need to implement this step because the SIFT-based point cloud is available in the dataset. Due to the large number of query images, we selected only 10 reference images per query. We then resized the width of each image to 864 and set the quantization size and the RANSAC threshold to 4 and 20, respectively. 

Table \ref{table:supp_cmu} provides a detailed breakdown of the results from the CMU benchmark, categorizing them based on different challenging conditions to assess the performance of localization methods. As demonstrated in Table \ref{table:supp_cmu}, DeViLoc consistently achieves over 90\% accuracy across all scenarios, surpassing other methods. 

\begin{figure*}[h]
  \centering
  \includegraphics[width=0.8\linewidth]{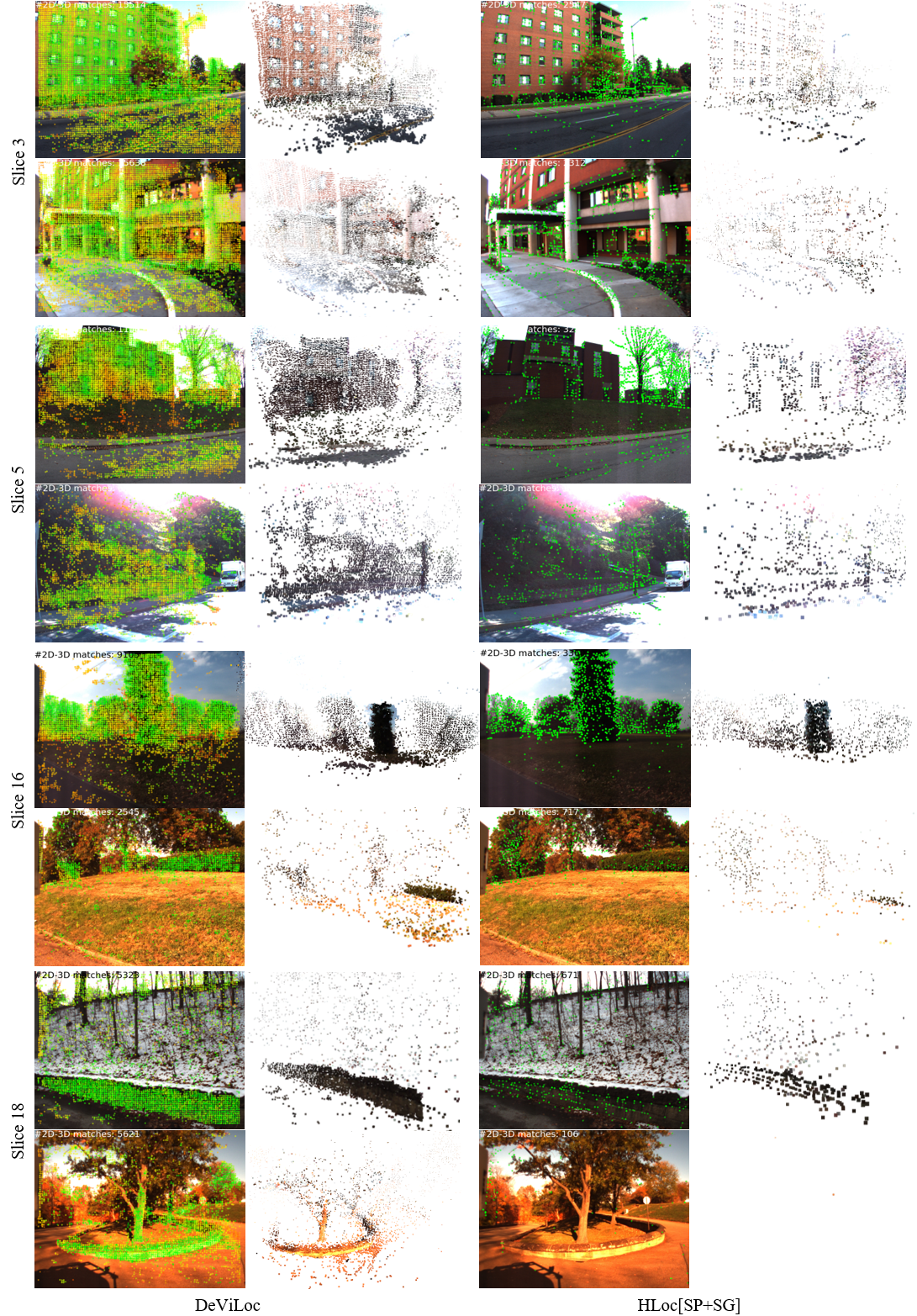}
  \caption{Qualitative comparison on the extended CMU seasons dataset. Compared to HLoc[SuperPoint+SuperGlue], DeViLoc can produce a higher number of accurate 2D-3D matches. Our method also estimates the matching confidence indicated by a color scale, from red (lowest confidence) to green (highest confidence).}
  \label{fig:supp_cmu}
\end{figure*}

Furthermore, Fig. \ref{fig:supp_cmu} offers a qualitative comparison between our method and HLoc[SP+SG]. The comparison reveals that DeViLoc generates numerous matches that effectively capture specific structures in the scenes. While our results may contain some noisy points, their overall impact is insignificant. In contrast, HLoc heavily relies on predetermined 3D points, leading to fewer 2D-3D matches in challenging conditions, as illustrated in Slices 16 and 18.

\section{Discussion}
\quad \textbf{Flexibility of DeViLoc}. While we utilized the detector-free image-matching model \cite{giang2023topicfm} to predict a substantial number of 2D-2D matches in the experiments, it is also possible to employ detector-based models for this step. In this situation, the runtime required for localization would be significantly reduced since the 2D-2D matching step is much more time-consuming than the proposed PIN and CPA modules (as demonstrated in Table 4 of the main paper). Due to the length and density of our paper, we have decided to leave this experiment to the audience after releasing the source code.

\textbf{Storage Demand}. Our database stores images and the 3D coordinates of point clouds, without the need of saving local features for all database images as done in HLoc. Moreover, during the experiments, we observed that our method maintains robust performance even when processing low-resolution images with a width of 864 pixels. Consequently, our localization system can reduce the storage demands by only saving low-resolution database images. However, the storage size of DeViLoc is still relatively large compared to methods that do not store local features or database images, such as NeuMap \cite{tang2023neumap} or GoMatch \cite{zhou2022geometry}. Nevertheless, these methods involve a tradeoff in localization accuracy. 
